\documentclass[11pt,a4paper]{article}
\usepackage[hyperref]{acl2020}
\usepackage{times}
\usepackage{latexsym}

\usepackage{microtype}

\usepackage{graphicx}
\usepackage{amsfonts}
\usepackage{amsmath}
\usepackage{amssymb}
\usepackage{pifont}
\usepackage{nicefrac}
\usepackage{multirow}
\usepackage{subfigure}
\usepackage[font=small]{caption}
\usepackage{soul}
\usepackage{xcolor}
\usepackage{booktabs}
\usepackage{comment}
\usepackage[linesnumbered,ruled,vlined]{algorithm2e}

\usepackage[utf8]{inputenc}
\usepackage[T1]{fontenc}

\usepackage{tablefootnote}

\definecolor{bblue}{HTML}{4F81BD}
\definecolor{rred}{HTML}{c4260b}
\definecolor{ggreen}{HTML}{098c1f}
\definecolor{ppurple}{HTML}{9F4C7C}
\definecolor{oorange}{HTML}{F79646}

\DeclareRobustCommand{\hlred}[1]{{\textcolor{rred}{#1}}}
\DeclareRobustCommand{\hbleu}[1]{{\textcolor{bblue}{#1}}}

\SetKwInOut{Input}{Input}

\usepackage{enumitem}
\setenumerate[1]{itemsep=0.30pt,partopsep=0pt,parsep=\parskip,topsep=5pt}
\setitemize[1]{itemsep=0.30pt,partopsep=00pt,parsep=\parskip,topsep=5pt}
\setdescription{itemsep=0.30pt,partopsep=0pt,parsep=\parskip,topsep=5pt}

\aclfinalcopy 

\newcommand\trobt{\textsc{ROBt}}
\newcommand\laln{\textsc{LaLn}}
\newcommand\lalt{\textsc{LaLt}}
\newcommand\matt{\textsc{MAtt}}
\newcommand\opus{{OPUS-100}}
\newcommand\bleufour{BLEU$_4$}
\newcommand\bleuall{BLEU$_{94}$}
\newcommand\bleuzero{BLEU$_{zero}$}
\newcommand\acczero{ACC$_{zero}$}

\usepackage{todonotes}

\title{Improving Massively Multilingual Neural Machine Translation and Zero-Shot Translation}

\author{Biao Zhang$^1$ \quad Philip Williams$^1$ \quad Ivan Titov$^{1,2}$ \quad Rico Sennrich$^{3,1}$ \bigskip\\
  $^1$School of Informatics, University of Edinburgh \\
  $^2$ILLC, University of Amsterdam \\
  $^3$Department of Computational Linguistics, University of Zurich \\
  \resizebox{\textwidth}{!}{\texttt{B.Zhang@ed.ac.uk, \{pwillia4,ititov\}@inf.ed.ac.uk, sennrich@cl.uzh.ch}}
  }

\date{}

\begin{document}
\maketitle
\begin{abstract}

Massively multilingual models for neural machine translation (NMT) are theoretically attractive, but often underperform bilingual models and deliver poor zero-shot translations. In this paper, we explore ways to improve them. 
We argue that multilingual NMT requires stronger modeling capacity to support language pairs 
with varying typological characteristics,
and overcome this bottleneck via language-specific components and deepening NMT architectures. We identify the off-target translation issue (i.e. translating into a wrong target language) as the major source of the inferior zero-shot performance, and propose random online backtranslation to enforce the translation of unseen training language pairs. 
Experiments on \opus{} (a novel multilingual dataset with 100 languages) show that our approach substantially narrows the performance gap with bilingual models in both one-to-many and many-to-many settings, and improves zero-shot performance by $\sim$10 BLEU, approaching  conventional pivot-based methods.\footnote{We release our code at \url{https://github.com/bzhangGo/zero}. We release the \opus{} dataset at \url{https://github.com/EdinburghNLP/opus-100-corpus}.}

\end{abstract}

\section{Introduction}

With the great success of neural machine translation (NMT) on bilingual datasets~\cite{DBLP:journals/corr/BahdanauCB14,NIPS2017_7181,barrault-etal-2019-findings}, there is renewed interest in multilingual translation where a single NMT model is optimized for the translation of multiple language pairs~\cite{firat-etal-2016-multi,johnson-etal-2017-googles,lu-etal-2018-neural,aharoni-etal-2019-massively}. Multilingual NMT eases model deployment and 
can encourage knowledge transfer among related language pairs~\cite{lakew-etal-2018-comparison,tan-etal-2019-multilingual}, improve low-resource translation~\cite{DBLP:journals/corr/HaNW16,DBLP:journals/corr/abs-1907-05019}, and enable zero-shot translation (i.e. direct translation between a language pair never seen in training)~\cite{firat-etal-2016-zero,johnson-etal-2017-googles,al-shedivat-parikh-2019-consistency,gu-etal-2019-improved}.

\begin{table}[t]
\centering
\small
\begin{tabular}{lp{5.5cm}}
\toprule
\multirow{2}{*}{Source} &  Jusqu'à ce qu'on trouve le moment clé, celui où tu as su que tu l'aimais. \\
\cmidrule{2-2}
\multirow{3}{*}{Reference} & Bis wir den unverkennbaren Moment gefunden haben, den Moment, wo du wusstest, du liebst ihn. \\
\cmidrule{2-2}
\multirow{2}{*}{Zero-Shot} & \hlred{Jusqu'à ce qu'on trouve le moment clé, celui où tu as su que tu l'aimais.} \\
\bottomrule
\multirow{2}{*}{Source} & Les États membres ont été consultés et ont approuvé cette proposition. \\
\cmidrule{2-2}
\multirow{2}{*}{Reference} & Die Mitgliedstaaten wurden konsultiert und sprachen sich für diesen Vorschlag aus. \\
\cmidrule{2-2}
\multirow{2}{*}{Zero-Shot} & \hlred{Les} \hbleu{Member States have been} consultedés \hbleu{and have approved this proposal.} \\
\bottomrule
\end{tabular}
\caption{\label{tb_off_target_issue} Illustration of the off-target translation issue with French$\to$German zero-shot translations with a multilingual NMT model. Our baseline multilingual NMT model often translates into the wrong language for zero-shot language pairs, such as \hlred{copying} the source sentence or translating into \hbleu{English} rather than German.}
\end{table}

Despite these potential benefits,  
multilingual NMT tends to underperform its bilingual counterparts~\cite{johnson-etal-2017-googles,DBLP:journals/corr/abs-1907-05019} and results in considerably worse translation performance when many languages are accommodated~\cite{aharoni-etal-2019-massively}. Since multilingual NMT must distribute its modeling capacity between different translation directions, we ascribe this deteriorated performance to the deficient capacity of single NMT models and seek solutions that are capable of overcoming this capacity bottleneck. We propose language-aware layer normalization and linear transformation to relax the representation constraint in multilingual NMT models. The linear transformation is inserted in-between the encoder and the decoder so as to facilitate the induction of language-specific translation correspondences. We also investigate
deep NMT architectures~\cite{wang-etal-2019-learning-deep,zhang-etal-2019-improving} aiming at further reducing the performance gap with bilingual methods.

Another pitfall of massively multilingual NMT is its poor zero-shot performance, particularly compared to pivot-based models. Without access to parallel training data for zero-shot language pairs, multilingual models easily fall into the trap of \textit{off-target translation} where a model ignores the given target information and translates into a wrong language as shown in Table \ref{tb_off_target_issue}. To avoid such a trap, we propose the random online backtranslation (\trobt) algorithm. \trobt{} finetunes a pretrained multilingual NMT model for unseen training language pairs with pseudo parallel batches generated by back-translating the target-side training data.\footnote{Note that backtranslation actually converts the zero-shot problem into a zero-resource problem. We follow previous work and continue referring to \textit{zero-shot} translation, even when using synthetic training data.} We perform backtranslation~\cite{sennrich-etal-2016-improving} into  randomly picked intermediate languages  to ensure good coverage of $\sim$10,000  zero-shot directions. Although backtranslation has been successfully applied to zero-shot translation ~\cite{firat-etal-2016-zero,gu-etal-2019-improved,2019arXiv190907342L}, whether it works in the massively multilingual set-up remained an open question and 
we investigate it in our work.

For experiments, we collect \opus{}, a massively multilingual dataset sampled from OPUS~\cite{Tiedemann2012}.
\opus{} consists of 55M English-centric sentence pairs covering 100 languages. 
As far as we know, no similar dataset is publicly available.\footnote{Previous studies~\cite{aharoni-etal-2019-massively,DBLP:journals/corr/abs-1907-05019} adopt in-house data which was not released.}
We have released \opus{} to facilitate future research.\footnote{\url{https://github.com/EdinburghNLP/opus-100-corpus}}
We adopt the Transformer model~\cite{NIPS2017_7181} and evaluate our approach under one-to-many and many-to-many translation settings. Our main findings are summarized as follows:
\begin{itemize}
    \item Increasing the capacity of multilingual NMT yields large improvements and narrows the performance gap with bilingual models. Low-resource translation benefits more from the increased capacity.
    \item Language-specific modeling and deep NMT architectures can slightly improve zero-shot translation, but fail to alleviate the off-target translation issue.
    \item Finetuning multilingual NMT with \trobt{} substantially reduces the proportion of off-target translations (by $\sim$50\%) and delivers an improvement of $\sim$10 BLEU in zero-shot settings, approaching the conventional pivot-based method. We show that finetuning with \trobt{} converges within a few thousand steps.
\end{itemize}

\section{Related Work}

Pioneering work on multilingual NMT began with multitask learning, which shared the encoder for one-to-many translation~\cite{dong-etal-2015-multi} or the attention mechanism for many-to-many translation~\cite{firat-etal-2016-multi}. These methods required a dedicated encoder or decoder for each language, limiting their scalability. 
By contrast, \citet{lee-etal-2017-fully} exploited character-level inputs and adopted a shared encoder for many-to-one translation. 
\citet{DBLP:journals/corr/HaNW16} and \citet{johnson-etal-2017-googles} further successfully trained a single NMT model for multilingual translation with a target language symbol guiding the translation direction. This approach serves as our baseline. Still, this paradigm forces different languages into one joint representation space, neglecting their linguistic diversity. Several subsequent studies have explored different strategies to mitigate this representation bottleneck, ranging from reorganizing parameter sharing~\cite{blackwood-etal-2018-multilingual,sachan-neubig-2018-parameter,lu-etal-2018-neural,wang-etal-2019-compact,vazquez-etal-2019-multilingual}, designing language-specific parameter generators~\cite{platanios-etal-2018-contextual}, decoupling multilingual word encodings~\cite{wang2018multilingual} to language clustering~\cite{tan-etal-2019-multilingual}. Our language-specific modeling continues in this direction, but with a special focus on broadening normalization layers and encoder outputs.

Multilingual NMT allows us to perform zero-shot translation, although the quality is not guaranteed~\cite{firat-etal-2016-zero,johnson-etal-2017-googles}. We observe that multilingual NMT often translates into the wrong target language on zero-shot directions (Table \ref{tb_off_target_issue}), resonating with the `missing ingredient problem'~\cite{DBLP:journals/corr/abs-1903-07091} and the spurious correlation issue~\cite{gu-etal-2019-improved}.  Approaches to improve zero-shot performance fall into two categories: 1) developing novel cross-lingual regularizers, such as the alignment regularizer~\cite{DBLP:journals/corr/abs-1903-07091} and the consistency regularizer~\cite{al-shedivat-parikh-2019-consistency}; and 2) generating artificial parallel data with backtranslation~\cite{firat-etal-2016-zero,gu-etal-2019-improved,2019arXiv190907342L} or pivot-based translation~\cite{currey-heafield-2019-zero}. The proposed \trobt{} algorithm belongs to the second category. In contrast to \citet{gu-etal-2019-improved} and \citet{2019arXiv190907342L}, however, we perform online backtranslation for each training step with randomly selected intermediate languages. \trobt{} avoids decoding the whole training set for each zero-shot language pair and can therefore scale to massively multilingual settings.

Our work belongs to a line of research on massively multilingual translation~\cite{aharoni-etal-2019-massively,DBLP:journals/corr/abs-1907-05019}. \citet{aharoni-etal-2019-massively} demonstrated the feasibility of massively multilingual NMT and reported encouraging results. We continue in this direction by developing approaches that improve both multilingual and zero-shot performance. Independently from our work, \citet{DBLP:journals/corr/abs-1907-05019} also find that increasing model capacity with deep architectures~\cite{wang-etal-2019-learning-deep,zhang-etal-2019-improving} substantially improves multilingual performance. A concurrent related work is~\cite{bapna-firat-2019-simple}, which introduces task-specific and lightweight adaptors for fast and scalable model adaptation. Compared to these adaptors, our language-aware layers are jointly trained with the whole NMT model from scratch without relying on any pretraining.

\section{Multilingual NMT}

We briefly review the multilingual approach~\cite{DBLP:journals/corr/HaNW16,johnson-etal-2017-googles} and the Transformer model~\cite{NIPS2017_7181}, which are used as our baseline. \citet{johnson-etal-2017-googles} rely on prepending tokens specifying the target language to each source sentence. In that way a single NMT model can be trained on the modified multilingual dataset and used to perform multilingual translation. Given a source sentence $\mathbf{x} $$ = $$ (x_1, x_2, \ldots, x_{|\mathbf{x}|})$, its target reference $\mathbf{y} $$ = $$ (y_1, y_2, \ldots, y_{|\mathbf{y}|})$ and the target language token  $t$\footnote{$t$ is in the form of ``{<2X>}'' where X is a language name, such as {<2EN>} meaning \textit{translating into English}.}, multilingual NMT translates under the encoder-decoder framework~\cite{DBLP:journals/corr/BahdanauCB14}:
\begin{align}
    \mathbf{H} & = \text{Encoder}([t, \mathbf{x}]), \label{eq_encoder} \\
    \mathbf{S} & = \text{Decoder}(\mathbf{y}, \mathbf{H}), \label{eq_decoder}
\end{align}
where $\mathbf{H} \in \mathbb{R}^{|\mathbf{x}|\times d}/\mathbf{S} \in \mathbb{R}^{|\mathbf{y}|\times d}$ denote the encoder/decoder output. $d$ is the model dimension.

We employ the Transformer~\cite{NIPS2017_7181} as the backbone NMT model due to its superior multilingual performance~\cite{lakew-etal-2018-comparison}. The encoder is a stack of $L=6$ identical layers, each  containing a self-attention sublayer and a point-wise feedforward sublayer. The decoder follows a similar structure, except for an extra cross-attention sublayer used to condition the decoder on the source sentence. Each sublayer is equipped with a residual connection~\cite{DBLP:journals/corr/HeZRS15}, followed by layer normalization~\cite[$\text{LN}(\cdot)$]{lei2016layer}:
\begin{equation}\label{eq_ln}
    \begin{split}
        \bar{\mathbf{a}} & = \text{LN}(\mathbf{a}\mid \mathbf{g}, \mathbf{b}) 
                           = \frac{\mathbf{a} - \mu}{\sigma} \odot \mathbf{g} + \mathbf{b},
    \end{split}
\end{equation}
where $\odot$ denotes element-wise multiplication, $\mu$ and $\sigma$ are the mean and standard deviation of the input vector $\mathbf{a} \in \mathbb{R}^d$, respectively. $\mathbf{g} \in \mathbb{R}^d$ and $\mathbf{b} \in \mathbb{R}^d$ are model parameters.
They control the sharpness and location of the regularized layer output $\bar{\mathbf{a}}$. Layer normalization has proven effective in accelerating model convergence~\cite{lei2016layer}.

\section{Approach}

Despite its success, multilingual NMT still suffers from 1) \textit{insufficient modeling capacity}, where including more languages results in reduction in translation quality~\cite{aharoni-etal-2019-massively}; and 2) \textit{off-target translation}, where models translate into a wrong target language on zero-shot directions~\cite{DBLP:journals/corr/abs-1903-07091}. These drawbacks become severe in massively multilingual settings and we explore approaches to alleviate them. 
We hypothesize that the vanilla Transformer has insufficient capacity and search for model-level strategies such as deepening Transformer and devising language-specific components. 
By contrast, we regard the lack of parallel data as the reason behind the off-target issue. We resort to data-level strategy by creating, in online fashion, artificial parallel training data for each zero-shot language pair in order to encourage its translation.

\paragraph{Deep Transformer} One natural way to improve the capacity is to increase model depth. Deeper neural models are often capable of inducing more generalizable (`abstract') representations and capturing more complex dependencies and have shown encouraging performance on bilingual translation~\cite{bapna-etal-2018-training,zhang-etal-2019-improving,wang-etal-2019-learning-deep}. We adopt the depth-scaled initialization method~\cite{zhang-etal-2019-improving} to train a deep Transformer for multilingual translation.

\paragraph{Language-aware Layer Normalization} Regardless of linguistic differences, layer normalization in multilingual NMT simply constrains all languages into one joint Gaussian space, which makes learning more difficult. We propose to relax this restriction by conditioning the normalization on the given target language token $t$ (\laln{} for short) as follows:
\begin{equation}\label{eq_la_ln}
    \bar{\mathbf{a}} = \text{LN}(\mathbf{a}\mid \mathbf{g}_t, \mathbf{b}_t).
\end{equation}
We apply this formula to all normalization layers, and leave the study of conditioning on source language information for the future. 

\paragraph{Language-aware Linear Transformation} Different language pairs have different translation correspondences or word alignments~\cite{Koehn:2010:SMT:1734086}. In addition to \laln{}, we introduce a target-language-aware linear transformation  (\lalt{} for short) between the encoder and the decoder to enhance the freedom of multilingual NMT in expressing flexible translation relationships. We adapt Eq. (\ref{eq_decoder}) as follows:
\begin{equation}\label{eq_la_linear}
    \mathbf{S} = \text{Decoder}(\mathbf{y}, \mathbf{H}\mathbf{W}_t),
\end{equation}
where $\mathbf{W}_t \in \mathbb{R}^{d\times d}$ denotes model parameters. Note that adding one more target language in \lalt{} brings in only one weight matrix.\footnote{We also attempted to factorize $\mathbf{W}_t$ into smaller matrices/vectors to reduce the number of parameters. Unfortunately, the final performance was rather disappointing.}
Compared to existing work~\cite{firat-etal-2016-zero,sachan-neubig-2018-parameter}, \lalt{} reaches a better trade-off between expressivity and scalability.

\begin{algorithm}[t]
\DontPrintSemicolon
\Input{Multilingual training data, $D$; \\
Pretrained multilingual model, $M$; \\
Maximum finetuning step, $N$; \\
Finetuning batch size, $B$; \\
Target language set, $\mathcal{T}$;
}
\KwOut{Zero-shot enabled model, $M$}
$i \gets 0$\;
\While{$i \leq N \land \text{not converged}$}{
   $\mathcal{B} \gets \text{sample batch from } D$\;
   \For{$k \gets 1$ \textbf{to} $B$}{
      $(\mathbf{x}_k, \mathbf{y}_k, t_k) \gets \mathcal{B}_k$\;
      $t_k^\prime \sim \text{Uniform}(\mathcal{T}) \text{ such that } t_k^\prime \ne t_k$\; 
      $\mathbf{x}_k^\prime \gets M([t_k^\prime, \mathbf{y}_k])$\; \label{alg_line_dec}
      \tcp*[l]{backtrans $t_k \rightarrow t_k^\prime$ to produce training example for $t_k^\prime \rightarrow t_k$}
      $\mathcal{B} \gets \mathcal{B} \cup (\mathbf{x}_k^\prime, \mathbf{y}_k, t_k)$\;
    }
    $\text{Optimize } M \text{ using } \mathcal{B}$\;
    $i \gets i + 1$\;
}
\Return{M}\;
\caption{\label{alg_robt}Algorithm for Random Online Backtranslation}
\end{algorithm}
\paragraph{Random Online Backtranslation} Prior studies on backtranslation for zero-shot translation decode the whole training set for each zero-shot language pair~\cite{gu-etal-2019-improved,2019arXiv190907342L}, and scalability to massively multilingual translation is questionable -- in our setting, the number of zero-shot translation directions is 9702.

We address scalability by performing online backtranslation paired with randomly sampled intermediate languages. Algorithm \ref{alg_robt} shows the detail of \trobt{}, where for each training instance $(\mathbf{x}_k, \mathbf{y}_k, t_k)$, we uniformly sample an intermediate language $t_k^\prime$ ($t_k \ne t_k^\prime$), back-translate $\mathbf{y}_k$ into $t_k^\prime$ to obtain $\mathbf{x}_k^\prime$, 
and train on the new instance $(\mathbf{x}_k^\prime, \mathbf{y}_k, t_k)$. Although $\mathbf{x}_k^\prime$ may be poor initially (translations are produced on-line by the model being trained), \trobt{} still benefits from the translation signal of $t_k^\prime \rightarrow t_k$. To reduce the computational cost, we implement batch-based greedy decoding for line \ref{alg_line_dec}.

\begin{table*}[t]
\centering
\small
\begin{tabular}{l|l|cr|rrr}
\toprule
{ID} & {Model Architecture} & $L$ & \#Param & \bleuall & {WR} & \bleufour{} \\
\midrule
1 & Transformer, Bilingual  & 6 & 106M & - & - & 20.90 \\
2 & Transformer, Bilingual  & 12 & 150M & - & - &  \textbf{22.75} \\
\midrule
3 & Transformer & 6 & 106M & 24.64 & \textit{ref} & 18.95 \\
4 & 3 + \matt{} & 6 & 99M & 23.81 & 20.2 & 17.95  \\
5 & 4 + \laln{} & 6 & 102M & 24.22 & 28.7 & 18.50 \\
6 & 4 + \lalt{} & 6 & 126M & 27.11 & 72.3 & 20.28 \\
7 & 4 + \laln{} + \lalt{} & 6 & 129M & 27.18 & 75.5 & 20.08 \\
\midrule
8 & 4 & 12 & 137M & 25.69 & 81.9 & 19.13 \\
9 & 7 & 12 & 169M & 28.04 & 91.5 & 19.93 \\
10 & 7 & 24 & 249M & \textbf{29.60} & \textbf{92.6} & 21.23 \\
\bottomrule
\end{tabular}
\caption{\label{tb_one_to_many} Test BLEU for one-to-many translation on \opus{} (100 languages). ``\textit{Bilingual}'': bilingual NMT, ``$L$'': model depth (for both encoder and decoder), ``\textit{\#Param}'': parameter number, ``\textit{WR}'': win ratio (\%) compared to \textit{ref} (\textcircled{3}), \matt{}: the merged attention~\cite{zhang-etal-2019-improving}. \laln{} and \lalt{} denote the proposed language-aware layer normalization and linear transformation, respectively. ``\textit{\bleuall{}}/\textit{\bleufour{}}'': average BLEU over all 94 translation directions in test set and En$\rightarrow$De/Zh/Br/Te, respectively. Higher BLEU and WR indicate better result. Best scores are highlighted in \textbf{bold}.}
\end{table*}

\section{\opus{}}
\label{sec:opus}

Recent work has scaled up multilingual NMT from a handful of languages to tens or hundreds, with many-to-many systems being capable of translation in thousands of directions.
Following \citet{aharoni-etal-2019-massively}, we created an English-centric dataset, meaning that all training pairs include English on either the source or target side.  Translation for any language pair that does not include English is zero-shot or must be pivoted through English.

We created \opus{} by sampling data from the OPUS collection~\cite{Tiedemann2012}.
\opus{} is at a similar scale to \citet{aharoni-etal-2019-massively}'s, with 100 languages (including English) on both sides and up to 1M training pairs for each language pair.
We selected the languages based on the volume of parallel data available in OPUS.

The OPUS collection is comprised of multiple corpora, ranging from movie subtitles to GNOME documentation to the Bible.
We did not curate the data or attempt to balance the representation of different domains, instead opting for the simplest approach of downloading all corpora for each language pair and concatenating them.
We randomly sampled up to 1M sentence pairs per language pair for training, as well as 2000 for validation and 2000 for testing.\footnote{For efficiency, we only use 200 sentences per language pair for validation in our multilingual experiments.}
To ensure that there was no overlap (at the monolingual sentence level) between the training and validation/test data, we applied a filter during sampling to exclude sentences that had already been sampled.
Note that this was done cross-lingually, so an English sentence in the Portuguese-English portion of the training data could not occur in the Hindi-English test set, for instance.

\opus{} contains approximately 55M sentence pairs.
Of the 99 language pairs, 44 have 1M sentence pairs of training data, 73 have at least 100k, and 95 have at least 10k.

To evaluate zero-shot translation, we also sampled 2000 sentence pairs of test data for each of the 15 pairings of Arabic, Chinese, Dutch, French, German, and Russian.
Filtering was used to exclude sentences already in \opus{}.

\begin{table*}[t]
\centering
\small
\begin{tabular}{l|l|cr|rrr|rrr}
\toprule
\multicolumn{1}{l}{\multirow{2}{*}{ID}} & \multicolumn{1}{l}{\multirow{2}{*}{Model Architecture}} & \multicolumn{1}{l}{\multirow{2}{*}{$L$}} & \multicolumn{1}{l}{\multirow{2}{*}{\#Param}} & \multicolumn{3}{c}{w/o \trobt{}} & \multicolumn{3}{|c}{w/ \trobt{}} \\
\cmidrule{5-10}
\multicolumn{1}{l}{}&\multicolumn{1}{l}{} &\multicolumn{1}{l}{} &\multicolumn{1}{l}{} & \bleuall{} & WR & \bleufour{} & \bleuall{} & WR & \bleufour{} \\
\midrule
1 & Transformer, Bilingual  & 6 & 110M & - & -  & \textbf{20.28} & - & - & -\\
\midrule
2 & Transformer             & 6 & 110M & 19.50 & \textit{ref} & 15.35 & 18.75 & 4.3 & 14.73 \\
3 & 2 + \matt{}    & 6 & 103M & 18.49 & 5.3 & 14.90 & 17.85 & 6.4 & 14.38 \\
4 & 3 + \laln{} + \lalt{} & 6 & 133M & 21.39 & 78.7 & 18.13 & 20.81 & 69.1 & 17.45 \\
\midrule
5 & 3 & 12 & 141M & 20.77 & 94.7 & 16.08 & 20.24 & 84.0 & 15.80 \\
6 & 4 & 12 & 173M & 22.86 & 97.9 & 19.25 & 22.39 & 97.9 & 18.23 \\
7 & 4 & 24 & 254M & \textbf{23.96} & \textbf{100.0} & 19.83 & 23.36 & 97.9 & 19.45 \\
\bottomrule
\end{tabular}
\caption{\label{tb_many_to_many_en_to_xx} English$\rightarrow$X test BLEU for many-to-many translation on \opus{} (100 languages). ``\textit{WR}'': win ratio (\%) compared to \textit{ref} (\textcircled{2} w/o \trobt{}).  \trobt{} denotes the proposed random online backtranslation method.} 
\end{table*}

\begin{table*}[t]
\centering
\small
\begin{tabular}{l|l|cr|rrr|rrr}
\toprule
\multicolumn{1}{l}{\multirow{2}{*}{ID}} & \multicolumn{1}{l}{\multirow{2}{*}{Model Architecture}} & \multicolumn{1}{l}{\multirow{2}{*}{$L$}} & \multicolumn{1}{l}{\multirow{2}{*}{\#Param}} & \multicolumn{3}{c}{w/o \trobt{}} & \multicolumn{3}{|c}{w/ \trobt{}} \\
\cmidrule{5-10}
\multicolumn{1}{l}{}&\multicolumn{1}{l}{} &\multicolumn{1}{l}{} &\multicolumn{1}{l}{} & \bleuall{} & WR & \bleufour{} & \bleuall{} & WR & \bleufour{} \\
\midrule
1 & Transformer, Bilingual  & 6 & 110M & - & - & 21.23 & - & - & - \\
\midrule
2 & Transformer             & 6 & 110M & 27.60 & \textit{ref} & 23.35 & 27.02 & 14.9 & 22.50 \\
3 & 2 + \matt{}    & 6 & 103M & 26.90 & 2.1 & 22.78 & 26.28 & 4.3 & 21.53 \\
4 & 3 + \laln{} + \lalt{} & 6 & 133M & 27.50 & 37.2 & 23.05 & 27.22 & 23.4 & 23.30 \\
\midrule
5 & 3  & 12 & 141M & 29.15 & \textbf{98.9} & 24.15 & 28.80 & 91.5 & 24.03 \\
6 & 4 & 12 & 173M & 29.49 & 97.9 & 24.53 & 29.54 & 96.8 & 25.43 \\
7 & 4 & 24 & 254M & \textbf{31.36} & \textbf{98.9} & 26.03 & 30.98 & 95.7 & \textbf{26.78} \\
\bottomrule
\end{tabular}
\caption{\label{tb_many_to_many_xx_to_en} X$\rightarrow$English test BLEU for many-to-many translation on \opus{} (100 languages). ``\textit{WR}'': win ratio (\%) compared to \textit{ref} (\textcircled{2} w/o \trobt{}).} 
\end{table*}
\section{Experiments}

\subsection{Setup}

We perform one-to-many (English-X) and many-to-many (English-X $\cup$ X-English) translation on \opus{} ($|\mathcal{T}|$ is 100). We apply byte pair encoding (BPE)~\cite{sennrich-etal-2016-neural,kudo-richardson-2018-sentencepiece} to handle multilingual words with a joint vocabulary size of 64k. We randomly shuffle the training set to mix instances of different language pairs. We adopt BLEU~\cite{papineni-etal-2002-bleu} for translation evaluation with the toolkit SacreBLEU~\cite{post-2018-call}\footnote{Signature: BLEU+case.mixed+numrefs.1+smooth.exp+\\tok.13a+version.1.4.1}. We employ the \textit{langdetect} library\footnote{\url{https://github.com/Mimino666/langdetect}} to detect the language of translations, and measure the translation-language accuracy for zero-shot cases. Rather than providing numbers for each language pair, we report average BLEU over all 94 language pairs with test sets (\bleuall{}). We also show the win ratio (WR), counting the proportion where our approach outperforms its baseline.

\begin{table*}[t]
\centering
\small
\begin{tabular}{l|l|cr|rrr|rrr}
\toprule
\multicolumn{1}{l}{\multirow{2}{*}{ID}} & \multicolumn{1}{l}{\multirow{2}{*}{Model Architecture}} & \multicolumn{1}{l}{\multirow{2}{*}{$L$}} & \multicolumn{1}{l}{\multirow{2}{*}{\#Param}} & \multicolumn{3}{c}{English$\rightarrow$X} & \multicolumn{3}{|c}{X$\rightarrow$English} \\
\cmidrule{5-10}
\multicolumn{1}{l}{}&\multicolumn{1}{l}{} &\multicolumn{1}{l}{} &\multicolumn{1}{l}{} & High & Med & Low & High & Med & Low \\
\midrule
1 & Transformer    & 6 & 110M & 20.69 & 20.82 & 15.18 & 26.99 & 28.60 & 27.49 \\
2 & 1 + \matt{}    & 6 & 103M & 19.70 & 19.77 & 14.17 & 26.32 & 27.81 & 26.84 \\
3 & 2 + \laln{} + \lalt{} & 6 & 133M & 21.07 & 22.88 & 19.99 & 27.03 & 28.60 & 26.97 \\
\midrule
4 & 2  & 12 & 141M & 21.67 & 22.17 & 16.95 & 28.39 & 30.24 & 29.26 \\
5 & 3 & 12 & 173M & 22.48 & 24.38 & 21.58 & 28.66 & 30.73 & 29.50 \\
6 & 3 & 24 & 254M & \textbf{23.69} & \textbf{25.61} & \textbf{22.24} & \textbf{30.29} & \textbf{32.58} & \textbf{31.90}  \\
\bottomrule
\end{tabular}
\caption{\label{tb_many_to_many_corpus} Test BLEU for High/Medium/Low (\textit{High/Med/Low}) resource language pairs in many-to-many setting on \opus{} (100 languages). We report average BLEU for each category.} 
\end{table*}

Apart from multilingual NMT, our baselines also involve bilingual NMT and pivot-based translation (only for zero-shot comparison). We select four typologically different target languages (German/De, Chinese/Zh, Breton/Br, Telugu/Te) with varied training data size for comparison to bilingual models as applying bilingual NMT to each language pair is resource-consuming. We report average BLEU over these four languages as \bleufour{}. We reuse the multilingual BPE vocabulary for bilingual NMT. 

We train all NMT models with the Transformer base settings (512/2048, 8 heads)~\cite{NIPS2017_7181}. We pair our approaches with the merged attention (\matt{})~\cite{zhang-etal-2019-improving} to reduce training time. Other details about model settings are in the Appendix.

\begin{table*}[t]
\centering
\small
\begin{tabular}{l|l|cr|rr|rr}
\toprule
\multicolumn{1}{l}{\multirow{2}{*}{ID}} & \multicolumn{1}{l}{\multirow{2}{*}{Model Architecture}} & \multicolumn{1}{l}{\multirow{2}{*}{$L$}} & \multicolumn{1}{l}{\multirow{2}{*}{\#Param}} & \multicolumn{2}{c}{w/o \trobt{}} & \multicolumn{2}{|c}{w/ \trobt{}} \\
\cmidrule{5-8}
\multicolumn{1}{l}{}&\multicolumn{1}{l}{} &\multicolumn{1}{l}{} &\multicolumn{1}{l}{} & \bleuzero{} & \acczero{} & \bleuzero{} & \acczero{} \\
\midrule
1 & Transformer, Pivot \& Bilingual  & 6 & 110M & 12.98 & 84.87 & - & - \\
\midrule
2 & Transformer              & 6 & 110M & 3.97 & 36.04 & 10.11 & 86.08 \\
3 & 2 + \matt{}    & 6 & 103M & 3.49 & 31.62 & 9.67 & 85.87 \\
4 & 3 + \laln{} + \lalt{} & 6 & 133M & 4.02 & 45.43 & 11.23 & 87.40 \\
\midrule
5 & 3 & 12 & 141M & 4.71 & 39.40 & 11.87 & 87.44 \\
6 & 4 & 12 & 173M & 5.41 & 51.40 & 12.62 & \textbf{87.99} \\
7 & 4 & 24 & 254M & 5.24 & 47.91 & 14.08 & 87.68 \\
\midrule
8 & 7 + Pivot & 24 & 254M & 14.71 & 84.81 & \textbf{14.78} & 85.09 \\
\bottomrule
\end{tabular}
\caption{\label{tb_many_to_many_zero_shot} Test BLEU and translation-language accuracy for zero-shot translation in many-to-many setting on \opus{} (100 languages). ``\textit{\bleuzero{}}/\textit{\acczero{}}'': average BLEU/accuracy over all zero-shot translation directions in test set, ``\textit{Pivot}'': the pivot-based translation that first translates one source sentence into English (X$\rightarrow$English NMT), and then into the target language (English$\rightarrow$X NMT). Lower accuracy indicates severe off-target translation. The average Pearson correlation coefficient between language accuracy and the corresponding BLEU is 0.93 (significant at $p<0.01$).}
\end{table*}

\subsection{Results on One-to-Many Translation}

Table \ref{tb_one_to_many} summarizes the results. The inferior performance of multilingual NMT (\textcircled{3}) against its bilingual counterpart (\textcircled{1}) reflects the capacity issue (-1.95 \bleufour{}).
Replacing the self-attention with \matt{} slightly deteriorates performance (-0.83 \bleuall{} \textcircled{3}$\rightarrow$\textcircled{4}); we still use \matt{} for more efficiently training deep models.

Our ablation study (\textcircled{4}-\textcircled{7}) shows that enriching the language awareness in multilingual NMT substantially alleviates this capacity problem. Relaxing the normalization constraints with \laln{} gains 0.41 \bleuall{} with 8.5\% WR (\textcircled{4}$\rightarrow$\textcircled{5}). Decoupling different translation relationships with \lalt{} delivers an improvement of 3.30 \bleuall{} and 52.1\% WR (\textcircled{4}$\rightarrow$\textcircled{6}). Combining \lalt{} and \laln{} demonstrates their complementarity (+3.37 \bleuall{} and +55.3\% WR, \textcircled{4}$\rightarrow$\textcircled{7}), significantly outperforming the multilingual baseline (+2.54 \bleuall{}, \textcircled{3}$\rightarrow$\textcircled{7}), albeit still behind the bilingual models (-0.82 \bleufour{}, \textcircled{1}$\rightarrow$\textcircled{7}).

Deepening the Transformer also improves the modeling capacity (+1.88 \bleuall{}, \textcircled{4}$\rightarrow$\textcircled{8}). Although deep Transformer performs worse than \laln{}+\lalt{} under a similar number of model parameters in terms of BLEU (-1.49 \bleuall{}, \textcircled{7}$\rightarrow$\textcircled{8}), it shows more consistent improvements across different language pairs (+6.4\% WR). 
We obtain better performance when integrating all approaches (\textcircled{9}). By increasing the model depth to 24 (\textcircled{10}), Transformer with our approach yields a score of 29.60 \bleuall{} and 21.23 \bleufour{}, beating the baseline (\textcircled{3}) on 92.6\% tasks and outperforming the base bilingual model (\textcircled{1}) by 0.33 \bleufour{}. Our approach significantly narrows the performance gap between multilingual NMT and bilingual NMT (20.90 \bleufour{} $\rightarrow$ 21.23 \bleufour{}, \textcircled{1}$\rightarrow$\textcircled{10}), although similarly deepening bilingual models surpasses our approach by 1.52 \bleufour{} (\textcircled{10}$\rightarrow$\textcircled{2}).

\subsection{Results on Many-to-Many Translation}\label{sec_result_many_to_many}

We train many-to-many NMT models on the concatenation of the one-to-many dataset (English$\rightarrow$X) and its reversed version (X$\rightarrow$English), and evaluate the zero-shot performance on X$\rightarrow$X language pairs. Table \ref{tb_many_to_many_en_to_xx} and Table \ref{tb_many_to_many_xx_to_en} show the translation results for English$\rightarrow$X and X$\rightarrow$English, respectively.\footnote{Note that the one-to-many training and test sets were not yet aggressively filtered for sentence overlap as described in Section~\ref{sec:opus}, so results in Table \ref{tb_one_to_many} and Table \ref{tb_many_to_many_en_to_xx} are not directly comparable.} We focus on the translation performance w/o \trobt{} in this subsection.

Compared to the one-to-many translation, the many-to-many translation must accommodate twice as many translation directions. We observe that many-to-many NMT models suffer more serious capacity issues on English$\rightarrow$X tasks (-4.93 \bleufour{}, \textcircled{1}$\rightarrow$\textcircled{2} in Table \ref{tb_many_to_many_en_to_xx} versus -1.95 \bleufour{} in Table \ref{tb_one_to_many}), where the deep Transformer with \laln{} + \lalt{} effectively reduces this gap to -0.45 \bleufour{} (\textcircled{1}$\rightarrow$\textcircled{7}, Table \ref{tb_many_to_many_en_to_xx}), resonating with our findings from Table \ref{tb_one_to_many}. By contrast, multilingual NMT benefits X$\rightarrow$English tasks considerably from the multitask learning alone, outperforming bilingual NMT by 2.13 \bleufour{} (\textcircled{1}$\rightarrow$\textcircled{2}, Table \ref{tb_many_to_many_xx_to_en}). Enhancing model capacity further enlarges this margin to +4.80 \bleufour{} (\textcircled{1}$\rightarrow$\textcircled{7}, Table \ref{tb_many_to_many_xx_to_en}).

We find that the overall quality of English$\rightarrow$X translation (19.50/23.96 \bleuall{}, \textcircled{2}/\textcircled{7}, Table \ref{tb_many_to_many_en_to_xx}) lags far behind that of its X$\rightarrow$English counterpart (27.60/31.36 \bleuall{}, \textcircled{2}/\textcircled{12}, Table \ref{tb_many_to_many_xx_to_en}), regardless of the modeling capacity. 
We ascribe this to the highly skewed training data distribution, where half of the training set uses English as the target. This strengthens the ability of the decoder to translate into English, and also encourages knowledge transfer for X$\rightarrow$English language pairs. 
\laln{} and \lalt{} show the largest benefit for English$\rightarrow$X (+2.9 \bleuall{}, \textcircled{3}$\rightarrow$\textcircled{4}, Table \ref{tb_many_to_many_en_to_xx}), and only a small benefit for X$\rightarrow$English (+0.6 \bleuall{}, \textcircled{3}$\rightarrow$\textcircled{4}, Table \ref{tb_many_to_many_xx_to_en}). This makes sense considering that \laln{} and \lalt{} are specific to the target language, so capacity is mainly increased for English$\rightarrow$X. Deepening the Transformer yields benefits in both directions (+2.57 \bleuall{} for English$\rightarrow$X, +3.86 \bleuall{} for X$\rightarrow$English; \textcircled{4}$\rightarrow$\textcircled{7}, Tables \ref{tb_many_to_many_en_to_xx} and \ref{tb_many_to_many_xx_to_en}).

\subsection{Effect of Training Corpus Size}

Our multilingual training data is distributed unevenly across different language pairs, which could affect the knowledge transfer delivered by language-aware modeling and deep Transformer in multilingual translation. We investigate this effect by grouping different language pairs in \opus{} into three categories according to their training data size: High ($\geq 0.9\text{M}$, 45), Low ($< 0.1\text{M}$, 18) and Medium (others, 31). Table \ref{tb_many_to_many_corpus} shows the results.

Language-aware modeling benefits low-resource language pairs the most on English$\rightarrow$X translation (+5.82 BLEU, Low versus +1.37/+3.11 BLEU, High/Med, \textcircled{2}$\rightarrow$\textcircled{3}), but has marginal impact on X$\rightarrow$English translation as analyzed in Section \ref{sec_result_many_to_many}. By contrast, deep Transformers yield similar benefits across different data scales (+2.38 average BLEU, English$\rightarrow$X and +2.31 average BLEU, X$\rightarrow$English, \textcircled{2}$\rightarrow$\textcircled{4}). We obtain the best performance by integrating both (\textcircled{1}$\rightarrow$\textcircled{6}) with a clear positive transfer to low-resource language pairs.

\subsection{Results on Zero-Shot Translation} 

Previous work shows that a well-trained multilingual model can do zero-shot  X$\rightarrow$Y translation directly~\cite{firat-etal-2016-zero,johnson-etal-2017-googles}. Our results in Table \ref{tb_many_to_many_zero_shot}  reveal that the translation quality is rather poor (3.97 \bleuzero{}, \textcircled{2} w/o \trobt{}) compared to the pivot-based bilingual baseline (12.98 \bleuzero{}, \textcircled{1}) under the massively multilingual setting~\cite{aharoni-etal-2019-massively}, although translations into different target languages show varied performance. The marginal gain by the deep Transformer with \laln{} + \lalt{} (+1.44 \bleuzero{}, \textcircled{2}$\rightarrow$\textcircled{6}, w/o \trobt{}) suggests that weak model capacity is not the major cause of this inferior performance. 

In a manual analysis on the zero-shot NMT outputs, we found many instances of off-target translation (Table \ref{tb_off_target_issue}). We use translation-language accuracy to measure the proportion of translations that are in the correct target language. Results in Table \ref{tb_many_to_many_zero_shot} show that there is a huge accuracy gap between the multilingual and the pivot-based method (-48.83\% \acczero{}, \textcircled{1}$\rightarrow$\textcircled{2}, w/o \trobt{}), from which we conclude that the off-target translation issue is one source of the poor zero-shot performance.

We apply \trobt{} to multilingual models by finetuning them for an extra 100k steps with the same batch size as for training. Table \ref{tb_many_to_many_zero_shot} shows that \trobt{} substantially improves \acczero{} by 35\%$\sim$50\%, reaching 85\%$\sim$87\% under different model settings. The multilingual Transformer with \trobt{} achieves a translation improvement of up to 10.11 \bleuzero{} (\textcircled{2} w/o \trobt{}$\rightarrow$\textcircled{7} w/ \trobt{}), outperforming the bilingual baseline by 1.1 \bleuzero{} (\textcircled{1} w/o \trobt{}$\rightarrow$\textcircled{7} w/ \trobt{}) and approaching the pivot-based multilingual baseline (-0.63 \bleuzero{}, \textcircled{8} w/o \trobt{}$\rightarrow$\textcircled{7} w/ \trobt{}).\footnote{Note that \trobt{} improves all zero-shot directions due to its randomness in sampling the intermediate languages. We do not bias \trobt{} to the given zero-shot test set.}  The strong Pearson correlation between the accuracy and BLEU (0.92 on average, significant at $p<0.01$) suggests that the improvement on the off-target translation issue explains the increased translation performance to a large extent.

\begin{figure}[t]
  \centering
    \includegraphics[scale=0.40]{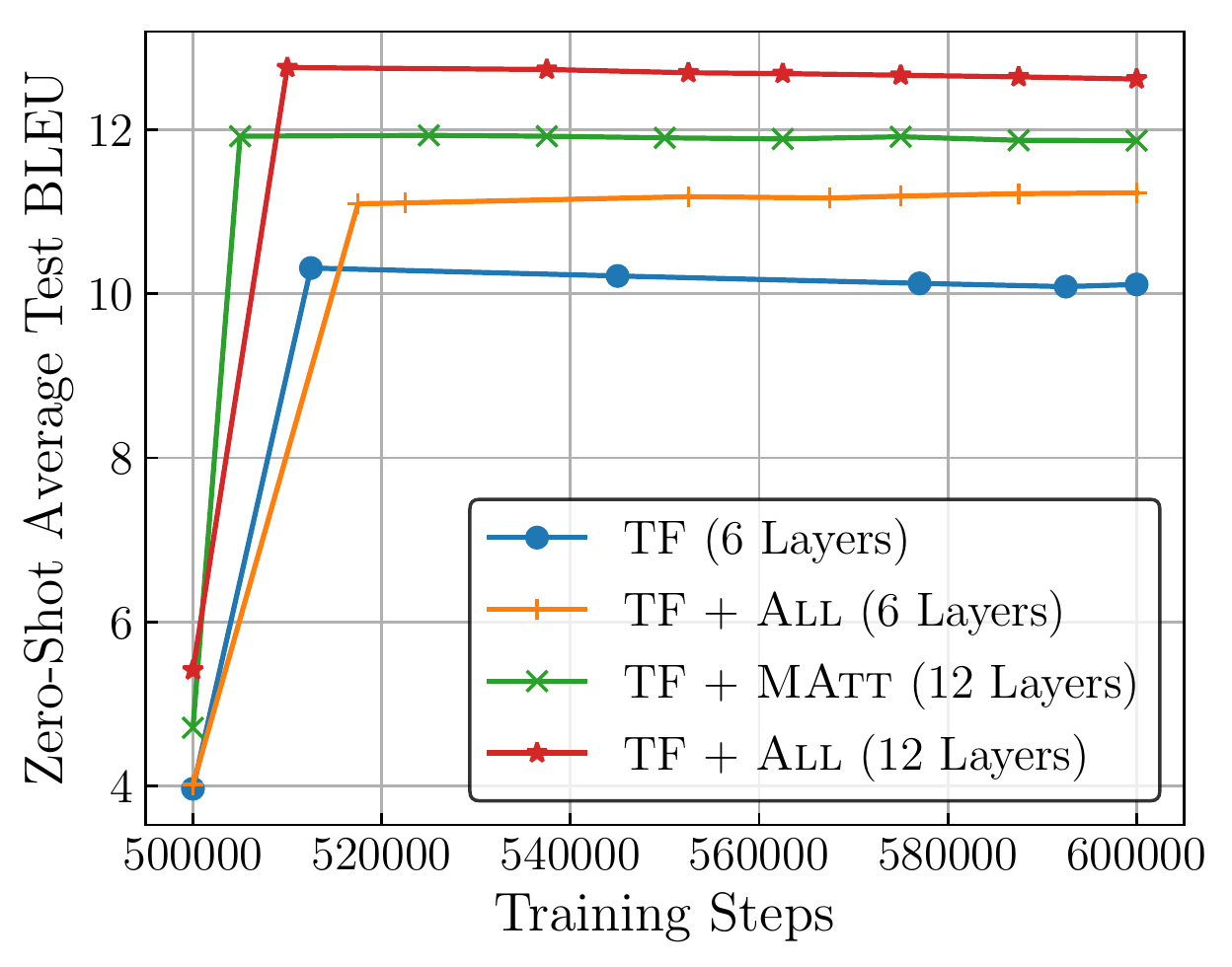}
  \caption{\label{fig_robt} Zero-shot average test BLEU for multilingual NMT models finetuned by \trobt{}. \textsc{All} = \matt{} + \laln{} + \lalt{}. Multilingual models with \trobt{} quickly converge on zero-shot directions.}
\end{figure}

\begin{table}[t]
\centering
\small
\begin{tabular}{cc}
\toprule
Setting & \bleuzero{} \\
\midrule
6-to-6 & 11.98 \\
100-to-100 & 11.23 \\
\bottomrule
\end{tabular}
\caption{\label{tb_langsize_zero_shot} Zero-short translation quality for \trobt{} under different settings. ``\textit{100-to-100}'': the setting used in the above experiments; we set $\mathcal{T}$ to all target languages. ``\textit{6-to-6}'': $\mathcal{T}$ only includes the zero-shot languages in the test set. We employ 6-layer Transformer with \laln{} and \lalt{} for experiments.}
\end{table}

Results in Table \ref{tb_many_to_many_en_to_xx} and \ref{tb_many_to_many_xx_to_en} show that \trobt{}'s success on zero-shot translation comes at the cost of sacrificing $\sim$0.50 \bleuall{} and $\sim$4\% WR on English$\rightarrow$X and X$\rightarrow$English translation. We also note that models with more capacity yield higher language accuracy (+7.78\%/+13.81\% \acczero{}, \textcircled{3}$\rightarrow$\textcircled{5}/\textcircled{3}$\rightarrow$\textcircled{4}, w/o \trobt{}) and deliver better zero-shot performance before (+1.22/+0.53 \bleuzero{}, \textcircled{3}$\rightarrow$\textcircled{5}/\textcircled{3}$\rightarrow$\textcircled{4}, w/o \trobt{}) and after \trobt{} (+2.20/+1.56 \bleuzero{}, \textcircled{3}$\rightarrow$\textcircled{5}/\textcircled{3}$\rightarrow$\textcircled{4}, w/ \trobt{}). 
In other words, increasing the modeling capacity benefits zero-shot translation and improves robustness.

\textit{Convergence of \trobt{}.} Unlike prior studies~\cite{gu-etal-2019-improved,2019arXiv190907342L}, we resort to an online method for backtranslation. The curve in Figure \ref{fig_robt} shows that \trobt{} is very effective, and takes only a few thousand steps to converge, suggesting that it is unnecessary to decode the whole training set for each zero-shot language pair. We leave it to future work to explore whether different back-translation strategies (other than greedy decoding) will deliver larger and continued benefits with \trobt{}.

\textit{Impact of $\mathcal{T}$ on \trobt{}.} \trobt{} heavily relies on $\mathcal{T}$, the set of target languages considered, to distribute the modeling capacity on zero-shot directions. To study its impact, we provide a comparison by constraining $\mathcal{T}$ to 6 languages in the zero-shot test set. Results in Table \ref{tb_langsize_zero_shot} show that the biased \trobt{} outperforms the baseline by 0.75 \bleuzero{}. By narrowing $\mathcal{T}$, more capacity is scheduled to the focused languages, which results in performance improvements. But the small scale of this improvement suggests that the number of zero-shot directions is not \trobt{}'s biggest bottleneck.

\section{Conclusion and Future Work}

This paper explores approaches to improve massively multilingual NMT, especially on zero-shot translation. We show that multilingual NMT suffers from weak capacity, and propose to enhance it by deepening the Transformer and devising language-aware neural models. We find that multilingual NMT often generates off-target translations on zero-shot directions, and propose to correct it with a random online backtranslation algorithm. 
We empirically demonstrate the feasibility of backtranslation in massively multilingual settings to allow for massively zero-shot translation for the first time. 
We release \opus{}, a multilingual dataset from OPUS including 100 languages with around 55M sentence pairs for future study. Our experiments on this dataset show that the proposed approaches substantially increase translation performance, narrowing the performance gap with bilingual NMT models and pivot-based methods. 

In the future, we will develop lightweight alternatives to \lalt{} to reduce the number of model parameters. We will also exploit novel strategies to break the upper bound of \trobt{} and obtain larger zero-shot improvements, such as generative modeling~\cite{zhang-etal-2016-variational-neural,su2018variational,Garca2020AMV,Zheng2020Mirror-Generative}.

\section*{Acknowledgments}

This project has received funding from the European Union’s Horizon 2020 Research and Innovation Programme under Grant Agreements 825460 (ELITR) and 825299 (GoURMET). 
This project has received support from Samsung Electronics Polska sp. z o.o. - Samsung R\&D Institute Poland.
Rico Sennrich acknowledges support of the Swiss National Science Foundation (MUTAMUR; no.\ 176727).

\bibliography{acl2020}

\begin{thebibliography}{43}
\expandafter\ifx\csname natexlab\endcsname\relax\def\natexlab#1{#1}\fi

\bibitem[{Aharoni et~al.(2019)Aharoni, Johnson, and
  Firat}]{aharoni-etal-2019-massively}
Roee Aharoni, Melvin Johnson, and Orhan Firat. 2019.
\newblock \href {https://doi.org/10.18653/v1/N19-1388} {Massively multilingual
  neural machine translation}.
\newblock In \emph{Proceedings of the 2019 Conference of the North {A}merican
  Chapter of the Association for Computational Linguistics: Human Language
  Technologies, Volume 1 (Long and Short Papers)}, pages 3874--3884,
  Minneapolis, Minnesota. Association for Computational Linguistics.

\bibitem[{Al-Shedivat and Parikh(2019)}]{al-shedivat-parikh-2019-consistency}
Maruan Al-Shedivat and Ankur Parikh. 2019.
\newblock \href {https://doi.org/10.18653/v1/N19-1121} {Consistency by
  agreement in zero-shot neural machine translation}.
\newblock In \emph{Proceedings of the 2019 Conference of the North {A}merican
  Chapter of the Association for Computational Linguistics: Human Language
  Technologies, Volume 1 (Long and Short Papers)}, pages 1184--1197,
  Minneapolis, Minnesota. Association for Computational Linguistics.

\bibitem[{Arivazhagan et~al.(2019{\natexlab{a}})Arivazhagan, Bapna, Firat,
  Aharoni, Johnson, and Macherey}]{DBLP:journals/corr/abs-1903-07091}
Naveen Arivazhagan, Ankur Bapna, Orhan Firat, Roee Aharoni, Melvin Johnson, and
  Wolfgang Macherey. 2019{\natexlab{a}}.
\newblock \href {http://arxiv.org/abs/1903.07091} {The missing ingredient in
  zero-shot neural machine translation}.
\newblock \emph{CoRR}, abs/1903.07091.

\bibitem[{Arivazhagan et~al.(2019{\natexlab{b}})Arivazhagan, Bapna, Firat,
  Lepikhin, Johnson, Krikun, Chen, Cao, Foster, Cherry, Macherey, Chen, and
  Wu}]{DBLP:journals/corr/abs-1907-05019}
Naveen Arivazhagan, Ankur Bapna, Orhan Firat, Dmitry Lepikhin, Melvin Johnson,
  Maxim Krikun, Mia~Xu Chen, Yuan Cao, George Foster, Colin Cherry, Wolfgang
  Macherey, Zhifeng Chen, and Yonghui Wu. 2019{\natexlab{b}}.
\newblock \href {http://arxiv.org/abs/1907.05019} {Massively multilingual
  neural machine translation in the wild: Findings and challenges}.
\newblock \emph{CoRR}, abs/1907.05019.

\bibitem[{Ba et~al.(2016)Ba, Kiros, and Hinton}]{lei2016layer}
Jimmy~Lei Ba, Jamie~Ryan Kiros, and Geoffrey~E Hinton. 2016.
\newblock Layer normalization.
\newblock \emph{arXiv preprint arXiv:1607.06450}.

\bibitem[{Bahdanau et~al.(2015)Bahdanau, Cho, and
  Bengio}]{DBLP:journals/corr/BahdanauCB14}
Dzmitry Bahdanau, Kyunghyun Cho, and Yoshua Bengio. 2015.
\newblock \href {http://arxiv.org/abs/1409.0473} {Neural machine translation by
  jointly learning to align and translate}.
\newblock In \emph{3rd International Conference on Learning Representations,
  {ICLR} 2015, San Diego, CA, USA, May 7-9, 2015, Conference Track
  Proceedings}.

\bibitem[{Bapna et~al.(2018)Bapna, Chen, Firat, Cao, and
  Wu}]{bapna-etal-2018-training}
Ankur Bapna, Mia Chen, Orhan Firat, Yuan Cao, and Yonghui Wu. 2018.
\newblock \href {https://doi.org/10.18653/v1/D18-1338} {Training deeper neural
  machine translation models with transparent attention}.
\newblock In \emph{Proceedings of the 2018 Conference on Empirical Methods in
  Natural Language Processing}, pages 3028--3033, Brussels, Belgium.
  Association for Computational Linguistics.

\bibitem[{Bapna and Firat(2019)}]{bapna-firat-2019-simple}
Ankur Bapna and Orhan Firat. 2019.
\newblock \href {https://doi.org/10.18653/v1/D19-1165} {Simple, scalable
  adaptation for neural machine translation}.
\newblock In \emph{Proceedings of the 2019 Conference on Empirical Methods in
  Natural Language Processing and the 9th International Joint Conference on
  Natural Language Processing (EMNLP-IJCNLP)}, pages 1538--1548, Hong Kong,
  China. Association for Computational Linguistics.

\bibitem[{Barrault et~al.(2019)Barrault, Bojar, Costa-juss{\`a}, Federmann,
  Fishel, Graham, Haddow, Huck, Koehn, Malmasi, Monz, M{\"u}ller, Pal, Post,
  and Zampieri}]{barrault-etal-2019-findings}
Lo{\"\i}c Barrault, Ond{\v{r}}ej Bojar, Marta~R. Costa-juss{\`a}, Christian
  Federmann, Mark Fishel, Yvette Graham, Barry Haddow, Matthias Huck, Philipp
  Koehn, Shervin Malmasi, Christof Monz, Mathias M{\"u}ller, Santanu Pal, Matt
  Post, and Marcos Zampieri. 2019.
\newblock \href {https://doi.org/10.18653/v1/W19-5301} {Findings of the 2019
  conference on machine translation ({WMT}19)}.
\newblock In \emph{Proceedings of the Fourth Conference on Machine Translation
  (Volume 2: Shared Task Papers, Day 1)}, pages 1--61, Florence, Italy.
  Association for Computational Linguistics.

\bibitem[{Blackwood et~al.(2018)Blackwood, Ballesteros, and
  Ward}]{blackwood-etal-2018-multilingual}
Graeme Blackwood, Miguel Ballesteros, and Todd Ward. 2018.
\newblock \href {https://www.aclweb.org/anthology/C18-1263} {Multilingual
  neural machine translation with task-specific attention}.
\newblock In \emph{Proceedings of the 27th International Conference on
  Computational Linguistics}, pages 3112--3122, Santa Fe, New Mexico, USA.
  Association for Computational Linguistics.

\bibitem[{Currey and Heafield(2019)}]{currey-heafield-2019-zero}
Anna Currey and Kenneth Heafield. 2019.
\newblock \href {https://doi.org/10.18653/v1/D19-5610} {Zero-resource neural
  machine translation with monolingual pivot data}.
\newblock In \emph{Proceedings of the 3rd Workshop on Neural Generation and
  Translation}, pages 99--107, Hong Kong. Association for Computational
  Linguistics.

\bibitem[{Dong et~al.(2015)Dong, Wu, He, Yu, and Wang}]{dong-etal-2015-multi}
Daxiang Dong, Hua Wu, Wei He, Dianhai Yu, and Haifeng Wang. 2015.
\newblock \href {https://doi.org/10.3115/v1/P15-1166} {Multi-task learning for
  multiple language translation}.
\newblock In \emph{Proceedings of the 53rd Annual Meeting of the Association
  for Computational Linguistics and the 7th International Joint Conference on
  Natural Language Processing (Volume 1: Long Papers)}, pages 1723--1732,
  Beijing, China. Association for Computational Linguistics.

\bibitem[{Firat et~al.(2016{\natexlab{a}})Firat, Cho, and
  Bengio}]{firat-etal-2016-multi}
Orhan Firat, Kyunghyun Cho, and Yoshua Bengio. 2016{\natexlab{a}}.
\newblock \href {https://doi.org/10.18653/v1/N16-1101} {Multi-way, multilingual
  neural machine translation with a shared attention mechanism}.
\newblock In \emph{Proceedings of the 2016 Conference of the North {A}merican
  Chapter of the Association for Computational Linguistics: Human Language
  Technologies}, pages 866--875, San Diego, California. Association for
  Computational Linguistics.

\bibitem[{Firat et~al.(2016{\natexlab{b}})Firat, Sankaran, Al-onaizan,
  Yarman~Vural, and Cho}]{firat-etal-2016-zero}
Orhan Firat, Baskaran Sankaran, Yaser Al-onaizan, Fatos~T. Yarman~Vural, and
  Kyunghyun Cho. 2016{\natexlab{b}}.
\newblock \href {https://doi.org/10.18653/v1/D16-1026} {Zero-resource
  translation with multi-lingual neural machine translation}.
\newblock In \emph{Proceedings of the 2016 Conference on Empirical Methods in
  Natural Language Processing}, pages 268--277, Austin, Texas. Association for
  Computational Linguistics.

\bibitem[{Garc{\'i}a et~al.(2020)Garc{\'i}a, For{\^e}t, Sellam, and
  Parikh}]{Garca2020AMV}
Xavier Garc{\'i}a, Pierre For{\^e}t, Thibault Sellam, and Ankur~P. Parikh.
  2020.
\newblock A multilingual view of unsupervised machine translation.
\newblock \emph{ArXiv}, abs/2002.02955.

\bibitem[{Gu et~al.(2019)Gu, Wang, Cho, and Li}]{gu-etal-2019-improved}
Jiatao Gu, Yong Wang, Kyunghyun Cho, and Victor~O.K. Li. 2019.
\newblock \href {https://doi.org/10.18653/v1/P19-1121} {Improved zero-shot
  neural machine translation via ignoring spurious correlations}.
\newblock In \emph{Proceedings of the 57th Annual Meeting of the Association
  for Computational Linguistics}, pages 1258--1268, Florence, Italy.
  Association for Computational Linguistics.

\bibitem[{Ha et~al.(2016)Ha, Niehues, and Waibel}]{DBLP:journals/corr/HaNW16}
Thanh-Le Ha, Jan Niehues, and Alexander Waibel. 2016.
\newblock Toward multilingual neural machine translation with universal encoder
  and decoder.
\newblock In \emph{Proceedings of the 13th International Workshop on Spoken
  Language Translation (IWSLT)}, Seattle, USA.

\bibitem[{He et~al.(2015)He, Zhang, Ren, and Sun}]{DBLP:journals/corr/HeZRS15}
Kaiming He, Xiangyu Zhang, Shaoqing Ren, and Jian Sun. 2015.
\newblock \href {http://arxiv.org/abs/1512.03385} {Deep residual learning for
  image recognition}.
\newblock \emph{CoRR}, abs/1512.03385.

\bibitem[{Johnson et~al.(2017)Johnson, Schuster, Le, Krikun, Wu, Chen, Thorat,
  Vi{\'e}gas, Wattenberg, Corrado, Hughes, and
  Dean}]{johnson-etal-2017-googles}
Melvin Johnson, Mike Schuster, Quoc~V. Le, Maxim Krikun, Yonghui Wu, Zhifeng
  Chen, Nikhil Thorat, Fernanda Vi{\'e}gas, Martin Wattenberg, Greg Corrado,
  Macduff Hughes, and Jeffrey Dean. 2017.
\newblock \href {https://doi.org/10.1162/tacl_a_00065} {{G}oogle{'}s
  multilingual neural machine translation system: Enabling zero-shot
  translation}.
\newblock \emph{Transactions of the Association for Computational Linguistics},
  5:339--351.

\bibitem[{Kingma and Ba(2015)}]{kingma2014adam}
Diederik~P Kingma and Jimmy Ba. 2015.
\newblock Adam: A method for stochastic optimization.
\newblock In \emph{International Conference on Learning Representations}.

\bibitem[{Koehn(2010)}]{Koehn:2010:SMT:1734086}
Philipp Koehn. 2010.
\newblock \emph{Statistical Machine Translation}, 1st edition.
\newblock Cambridge University Press, New York, NY, USA.

\bibitem[{Kudo and Richardson(2018)}]{kudo-richardson-2018-sentencepiece}
Taku Kudo and John Richardson. 2018.
\newblock \href {https://doi.org/10.18653/v1/D18-2012} {{S}entence{P}iece: A
  simple and language independent subword tokenizer and detokenizer for neural
  text processing}.
\newblock In \emph{Proceedings of the 2018 Conference on Empirical Methods in
  Natural Language Processing: System Demonstrations}, pages 66--71, Brussels,
  Belgium. Association for Computational Linguistics.

\bibitem[{{Lakew} et~al.(2019){Lakew}, {Federico}, {Negri}, and
  {Turchi}}]{2019arXiv190907342L}
Surafel~M. {Lakew}, Marcello {Federico}, Matteo {Negri}, and Marco {Turchi}.
  2019.
\newblock \href {http://arxiv.org/abs/1909.07342} {{Multilingual Neural Machine
  Translation for Zero-Resource Languages}}.
\newblock \emph{arXiv e-prints}, page arXiv:1909.07342.

\bibitem[{Lakew et~al.(2018)Lakew, Cettolo, and
  Federico}]{lakew-etal-2018-comparison}
Surafel~Melaku Lakew, Mauro Cettolo, and Marcello Federico. 2018.
\newblock \href {https://www.aclweb.org/anthology/C18-1054} {A comparison of
  transformer and recurrent neural networks on multilingual neural machine
  translation}.
\newblock In \emph{Proceedings of the 27th International Conference on
  Computational Linguistics}, pages 641--652, Santa Fe, New Mexico, USA.
  Association for Computational Linguistics.

\bibitem[{Lee et~al.(2017)Lee, Cho, and Hofmann}]{lee-etal-2017-fully}
Jason Lee, Kyunghyun Cho, and Thomas Hofmann. 2017.
\newblock \href {https://doi.org/10.1162/tacl_a_00067} {Fully character-level
  neural machine translation without explicit segmentation}.
\newblock \emph{Transactions of the Association for Computational Linguistics},
  5:365--378.

\bibitem[{Lu et~al.(2018)Lu, Keung, Ladhak, Bhardwaj, Zhang, and
  Sun}]{lu-etal-2018-neural}
Yichao Lu, Phillip Keung, Faisal Ladhak, Vikas Bhardwaj, Shaonan Zhang, and
  Jason Sun. 2018.
\newblock \href {https://doi.org/10.18653/v1/W18-6309} {A neural interlingua
  for multilingual machine translation}.
\newblock In \emph{Proceedings of the Third Conference on Machine Translation:
  Research Papers}, pages 84--92, Brussels, Belgium. Association for
  Computational Linguistics.

\bibitem[{Papineni et~al.(2002)Papineni, Roukos, Ward, and
  Zhu}]{papineni-etal-2002-bleu}
Kishore Papineni, Salim Roukos, Todd Ward, and Wei-Jing Zhu. 2002.
\newblock \href {https://doi.org/10.3115/1073083.1073135} {{B}leu: a method for
  automatic evaluation of machine translation}.
\newblock In \emph{Proceedings of the 40th Annual Meeting of the Association
  for Computational Linguistics}, pages 311--318, Philadelphia, Pennsylvania,
  USA. Association for Computational Linguistics.

\bibitem[{Platanios et~al.(2018)Platanios, Sachan, Neubig, and
  Mitchell}]{platanios-etal-2018-contextual}
Emmanouil~Antonios Platanios, Mrinmaya Sachan, Graham Neubig, and Tom Mitchell.
  2018.
\newblock \href {https://doi.org/10.18653/v1/D18-1039} {Contextual parameter
  generation for universal neural machine translation}.
\newblock In \emph{Proceedings of the 2018 Conference on Empirical Methods in
  Natural Language Processing}, pages 425--435, Brussels, Belgium. Association
  for Computational Linguistics.

\bibitem[{Post(2018)}]{post-2018-call}
Matt Post. 2018.
\newblock \href {https://www.aclweb.org/anthology/W18-6319} {A call for clarity
  in reporting {BLEU} scores}.
\newblock In \emph{Proceedings of the Third Conference on Machine Translation:
  Research Papers}, pages 186--191, Belgium, Brussels. Association for
  Computational Linguistics.

\bibitem[{Sachan and Neubig(2018)}]{sachan-neubig-2018-parameter}
Devendra Sachan and Graham Neubig. 2018.
\newblock \href {https://doi.org/10.18653/v1/W18-6327} {Parameter sharing
  methods for multilingual self-attentional translation models}.
\newblock In \emph{Proceedings of the Third Conference on Machine Translation:
  Research Papers}, pages 261--271, Brussels, Belgium. Association for
  Computational Linguistics.

\bibitem[{Sennrich et~al.(2016{\natexlab{a}})Sennrich, Haddow, and
  Birch}]{sennrich-etal-2016-improving}
Rico Sennrich, Barry Haddow, and Alexandra Birch. 2016{\natexlab{a}}.
\newblock \href {https://doi.org/10.18653/v1/P16-1009} {Improving neural
  machine translation models with monolingual data}.
\newblock In \emph{Proceedings of the 54th Annual Meeting of the Association
  for Computational Linguistics (Volume 1: Long Papers)}, pages 86--96, Berlin,
  Germany. Association for Computational Linguistics.

\bibitem[{Sennrich et~al.(2016{\natexlab{b}})Sennrich, Haddow, and
  Birch}]{sennrich-etal-2016-neural}
Rico Sennrich, Barry Haddow, and Alexandra Birch. 2016{\natexlab{b}}.
\newblock \href {https://doi.org/10.18653/v1/P16-1162} {Neural machine
  translation of rare words with subword units}.
\newblock In \emph{Proceedings of the 54th Annual Meeting of the Association
  for Computational Linguistics (Volume 1: Long Papers)}, pages 1715--1725,
  Berlin, Germany. Association for Computational Linguistics.

\bibitem[{Su et~al.(2018)Su, Wu, Xiong, Lu, Han, and Zhang}]{su2018variational}
Jinsong Su, Shan Wu, Deyi Xiong, Yaojie Lu, Xianpei Han, and Biao Zhang. 2018.
\newblock Variational recurrent neural machine translation.
\newblock In \emph{Thirty-Second AAAI Conference on Artificial Intelligence}.

\bibitem[{Tan et~al.(2019)Tan, Chen, He, Xia, QIN, and
  Liu}]{tan-etal-2019-multilingual}
Xu~Tan, Jiale Chen, Di~He, Yingce Xia, Tao QIN, and Tie-Yan Liu. 2019.
\newblock \href {https://doi.org/10.18653/v1/D19-1089} {Multilingual neural
  machine translation with language clustering}.
\newblock In \emph{Proceedings of the 2019 Conference on Empirical Methods in
  Natural Language Processing and the 9th International Joint Conference on
  Natural Language Processing (EMNLP-IJCNLP)}, pages 963--973, Hong Kong,
  China. Association for Computational Linguistics.

\bibitem[{Tiedemann(2012)}]{Tiedemann2012}
Jörg Tiedemann. 2012.
\newblock Parallel data, tools and interfaces in opus.
\newblock In \emph{Proceedings of the Eight International Conference on
  Language Resources and Evaluation (LREC'12)}, Istanbul, Turkey. European
  Language Resources Association (ELRA).

\bibitem[{Vaswani et~al.(2017)Vaswani, Shazeer, Parmar, Uszkoreit, Jones,
  Gomez, Kaiser, and Polosukhin}]{NIPS2017_7181}
Ashish Vaswani, Noam Shazeer, Niki Parmar, Jakob Uszkoreit, Llion Jones,
  Aidan~N Gomez, \L~ukasz Kaiser, and Illia Polosukhin. 2017.
\newblock \href
  {http://papers.nips.cc/paper/7181-attention-is-all-you-need.pdf} {Attention
  is all you need}.
\newblock In I.~Guyon, U.~V. Luxburg, S.~Bengio, H.~Wallach, R.~Fergus,
  S.~Vishwanathan, and R.~Garnett, editors, \emph{Advances in Neural
  Information Processing Systems 30}, pages 5998--6008. Curran Associates, Inc.

\bibitem[{V{\'a}zquez et~al.(2019)V{\'a}zquez, Raganato, Tiedemann, and
  Creutz}]{vazquez-etal-2019-multilingual}
Ra{\'u}l V{\'a}zquez, Alessandro Raganato, J{\"o}rg Tiedemann, and Mathias
  Creutz. 2019.
\newblock \href {https://doi.org/10.18653/v1/W19-4305} {Multilingual {NMT} with
  a language-independent attention bridge}.
\newblock In \emph{Proceedings of the 4th Workshop on Representation Learning
  for NLP (RepL4NLP-2019)}, pages 33--39, Florence, Italy. Association for
  Computational Linguistics.

\bibitem[{Wang et~al.(2019{\natexlab{a}})Wang, Li, Xiao, Zhu, Li, Wong, and
  Chao}]{wang-etal-2019-learning-deep}
Qiang Wang, Bei Li, Tong Xiao, Jingbo Zhu, Changliang Li, Derek~F. Wong, and
  Lidia~S. Chao. 2019{\natexlab{a}}.
\newblock \href {https://doi.org/10.18653/v1/P19-1176} {Learning deep
  transformer models for machine translation}.
\newblock In \emph{Proceedings of the 57th Annual Meeting of the Association
  for Computational Linguistics}, pages 1810--1822, Florence, Italy.
  Association for Computational Linguistics.

\bibitem[{Wang et~al.(2019{\natexlab{b}})Wang, Pham, Arthur, and
  Neubig}]{wang2018multilingual}
Xinyi Wang, Hieu Pham, Philip Arthur, and Graham Neubig. 2019{\natexlab{b}}.
\newblock \href {https://openreview.net/forum?id=Skeke3C5Fm} {Multilingual
  neural machine translation with soft decoupled encoding}.
\newblock In \emph{International Conference on Learning Representations}.

\bibitem[{Wang et~al.(2019{\natexlab{c}})Wang, Zhou, Zhang, Zhai, Xu, and
  Zong}]{wang-etal-2019-compact}
Yining Wang, Long Zhou, Jiajun Zhang, Feifei Zhai, Jingfang Xu, and Chengqing
  Zong. 2019{\natexlab{c}}.
\newblock \href {https://doi.org/10.18653/v1/P19-1117} {A compact and
  language-sensitive multilingual translation method}.
\newblock In \emph{Proceedings of the 57th Annual Meeting of the Association
  for Computational Linguistics}, pages 1213--1223, Florence, Italy.
  Association for Computational Linguistics.

\bibitem[{Zhang et~al.(2019)Zhang, Titov, and
  Sennrich}]{zhang-etal-2019-improving}
Biao Zhang, Ivan Titov, and Rico Sennrich. 2019.
\newblock \href {https://doi.org/10.18653/v1/D19-1083} {Improving deep
  transformer with depth-scaled initialization and merged attention}.
\newblock In \emph{Proceedings of the 2019 Conference on Empirical Methods in
  Natural Language Processing and the 9th International Joint Conference on
  Natural Language Processing (EMNLP-IJCNLP)}, pages 898--909, Hong Kong,
  China. Association for Computational Linguistics.

\bibitem[{Zhang et~al.(2016)Zhang, Xiong, Su, Duan, and
  Zhang}]{zhang-etal-2016-variational-neural}
Biao Zhang, Deyi Xiong, Jinsong Su, Hong Duan, and Min Zhang. 2016.
\newblock \href {https://doi.org/10.18653/v1/D16-1050} {Variational neural
  machine translation}.
\newblock In \emph{Proceedings of the 2016 Conference on Empirical Methods in
  Natural Language Processing}, pages 521--530, Austin, Texas. Association for
  Computational Linguistics.

\bibitem[{Zheng et~al.(2020)Zheng, Zhou, Huang, Li, Dai, and
  Chen}]{Zheng2020Mirror-Generative}
Zaixiang Zheng, Hao Zhou, Shujian Huang, Lei Li, Xin-Yu Dai, and Jiajun Chen.
  2020.
\newblock \href {https://openreview.net/forum?id=HkxQRTNYPH} {Mirror-generative
  neural machine translation}.
\newblock In \emph{International Conference on Learning Representations}.

\end{thebibliography}
\bibliographystyle{acl_natbib}

\clearpage

\appendix

\section{\opus{}: The OPUS Multilingual Dataset}
\label{app:dataset}

Table~\ref{tab:en-centric} lists the languages (other than English) and numbers of sentence pairs in the English-centric multilingual dataset.

\begin{table*}[h!]
\centering
\small
\caption{Numbers of training, validation, and test sentence pairs in the English-centric multilingual dataset.}
\label{tab:en-centric}
\begin{tabular}{llrrrp{1cm}llrrr}
  \multicolumn{2}{l}{Language} & Train & Valid & Test & & \multicolumn{2}{l}{Language} & Train & Valid & Test \\
  \cline{1-5} \cline{7-11}
  af & Afrikaans & 275512 & 2000 & 2000 & & lv & Latvian & 1000000 & 2000 & 2000 \\
  am & Amharic & 89027 & 2000 & 2000 & & mg & Malagasy & 590771 & 2000 & 2000 \\
  an & Aragonese & 6961 & 0 & 0 & & mk & Macedonian & 1000000 & 2000 & 2000 \\
  ar & Arabic & 1000000 & 2000 & 2000 & & ml & Malayalam & 822746 & 2000 & 2000 \\
  as & Assamese & 138479 & 2000 & 2000 & & mn & Mongolian & 4294 & 0 & 0 \\
  az & Azerbaijani & 262089 & 2000 & 2000 & & mr & Marathi & 27007 & 2000 & 2000 \\
  be & Belarusian & 67312 & 2000 & 2000 & & ms & Malay & 1000000 & 2000 & 2000 \\
  bg & Bulgarian & 1000000 & 2000 & 2000 & & mt & Maltese & 1000000 & 2000 & 2000 \\
  bn & Bengali & 1000000 & 2000 & 2000 & & my & Burmese & 24594 & 2000 & 2000 \\
  br & Breton & 153447 & 2000 & 2000 & & nb & Norwegian Bokmål & 142906 & 2000 & 2000 \\
  bs & Bosnian & 1000000 & 2000 & 2000 & & ne & Nepali & 406381 & 2000 & 2000 \\
  ca & Catalan & 1000000 & 2000 & 2000 & & nl & Dutch & 1000000 & 2000 & 2000 \\
  cs & Czech & 1000000 & 2000 & 2000 & & nn & Norwegian Nynorsk & 486055 & 2000 & 2000 \\
  cy & Welsh & 289521 & 2000 & 2000 & & no & Norwegian & 1000000 & 2000 & 2000 \\
  da & Danish & 1000000 & 2000 & 2000 & & oc & Occitan & 35791 & 2000 & 2000 \\
  de & German & 1000000 & 2000 & 2000 & & or & Oriya & 14273 & 1317 & 1318 \\
  dz & Dzongkha & 624 & 0 & 0 & & pa & Panjabi & 107296 & 2000 & 2000 \\
  el & Greek & 1000000 & 2000 & 2000 & & pl & Polish & 1000000 & 2000 & 2000 \\
  eo & Esperanto & 337106 & 2000 & 2000 & & ps & Pashto & 79127 & 2000 & 2000 \\
  es & Spanish & 1000000 & 2000 & 2000 & & pt & Portuguese & 1000000 & 2000 & 2000 \\
  et & Estonian & 1000000 & 2000 & 2000 & & ro & Romanian & 1000000 & 2000 & 2000 \\
  eu & Basque & 1000000 & 2000 & 2000 & & ru & Russian & 1000000 & 2000 & 2000 \\
  fa & Persian & 1000000 & 2000 & 2000 & & rw & Kinyarwanda & 173823 & 2000 & 2000 \\
  fi & Finnish & 1000000 & 2000 & 2000 & & se & Northern Sami & 35907 & 2000 & 2000 \\
  fr & French & 1000000 & 2000 & 2000 & & sh & Serbo-Croatian & 267211 & 2000 & 2000 \\
  fy & Western Frisian & 54342 & 2000 & 2000 & & si & Sinhala & 979109 & 2000 & 2000 \\
  ga & Irish & 289524 & 2000 & 2000 & & sk & Slovak & 1000000 & 2000 & 2000 \\
  gd & Gaelic & 16316 & 1605 & 1606 & & sl & Slovenian & 1000000 & 2000 & 2000 \\
  gl & Galician & 515344 & 2000 & 2000 & & sq & Albanian & 1000000 & 2000 & 2000 \\
  gu & Gujarati & 318306 & 2000 & 2000 & & sr & Serbian & 1000000 & 2000 & 2000 \\
  ha & Hausa & 97983 & 2000 & 2000 & & sv & Swedish & 1000000 & 2000 & 2000 \\
  he & Hebrew & 1000000 & 2000 & 2000 & & ta & Tamil & 227014 & 2000 & 2000 \\
  hi & Hindi & 534319 & 2000 & 2000 & & te & Telugu & 64352 & 2000 & 2000 \\
  hr & Croatian & 1000000 & 2000 & 2000 & & tg & Tajik & 193882 & 2000 & 2000 \\
  hu & Hungarian & 1000000 & 2000 & 2000 & & th & Thai & 1000000 & 2000 & 2000 \\
  hy & Armenian & 7059 & 0 & 0 & & tk & Turkmen & 13110 & 1852 & 1852 \\
  id & Indonesian & 1000000 & 2000 & 2000 & & tr & Turkish & 1000000 & 2000 & 2000 \\
  ig & Igbo & 18415 & 1843 & 1843 & & tt & Tatar & 100843 & 2000 & 2000 \\
  is & Icelandic & 1000000 & 2000 & 2000 & & ug & Uighur & 72170 & 2000 & 2000 \\
  it & Italian & 1000000 & 2000 & 2000 & & uk & Ukrainian & 1000000 & 2000 & 2000 \\
  ja & Japanese & 1000000 & 2000 & 2000 & & ur & Urdu & 753913 & 2000 & 2000 \\
  ka & Georgian & 377306 & 2000 & 2000 & & uz & Uzbek & 173157 & 2000 & 2000 \\
  kk & Kazakh & 79927 & 2000 & 2000 & & vi & Vietnamese & 1000000 & 2000 & 2000 \\
  km & Central Khmer & 111483 & 2000 & 2000 & & wa & Walloon & 104496 & 2000 & 2000 \\
  kn & Kannada & 14537 & 917 & 918 & & xh & Xhosa & 439671 & 2000 & 2000 \\
  ko & Korean & 1000000 & 2000 & 2000 & & yi & Yiddish & 15010 & 2000 & 2000 \\
  ku & Kurdish & 144844 & 2000 & 2000 & & yo & Yoruba & 10375 & 0 & 0 \\
  ky & Kyrgyz & 27215 & 2000 & 2000 & & zh & Chinese & 1000000 & 2000 & 2000 \\
  li & Limburgan & 25535 & 2000 & 2000 & & zu & Zulu & 38616 & 2000 & 2000 \\
  lt & Lithuanian & 1000000 & 2000 & 2000 & & & & & &
\end{tabular}
\end{table*}

\section{Model Settings}
We optimize model parameters using Adam ($\beta_1=0.9, \beta_2=0.98$)~\cite{kingma2014adam} with label smoothing of 0.1 and scheduled learning rate (warmup step 4k). We set the initial learning rate to 1.0 for bilingual models, but use 0.5 for multilingual models in order to stabilize training. We apply dropout to residual layers and attention weights, with a rate of 0.1/0.1 for 6-layer Transformer models and 0.3/0.2 for deeper ones. We group sentence pairs of roughly 50k target tokens into one training/finetuning batch, except for bilingual models where 25k target tokens are used. We train multilingual and bilingual models for 500k and 100k steps, respectively. We average the last 5 checkpoints for evaluation, and employ beam search for decoding with a beam size of 4 and length penalty of 0.6.


\end{document}